\documentclass{article}
\usepackage[preprint]{neurips_2025}

\usepackage[utf8]{inputenc} %
\usepackage[T1]{fontenc}    %
\usepackage{hyperref}       %
\usepackage{url}            %
\usepackage{booktabs}       %
\usepackage{amsfonts}       %
\usepackage{amsmath}  
\usepackage{nicefrac}       %
\usepackage{microtype}      %
\usepackage{xcolor}         %
\usepackage{pifont}
\usepackage{pgfplots}
\pgfplotsset{compat=1.18} 
\usepackage{makecell}
\usepackage{fancyvrb}
\usepackage{fvextra} %
\usepackage{listings}

\definecolor{diffgreen}{HTML}{228B22} %

\DefineVerbatimEnvironment{Prompt}{Verbatim}{
  breaklines=true,
  breakanywhere=true, 
  breaksymbolleft={}, breaksymbolright={}, breaksymbolindent=0pt,
  formatcom=\ttfamily\small
}

\lstdefinestyle{diffstyle}{
    language=[LaTeX]tex,
    frame=tb, %
    tabsize=2,
    basicstyle=\ttfamily\small,
    moredelim=[il][\bfseries\color{diffgreen}]{+},
    showstringspaces=false,
    breaklines=true,
    prebreak=\raisebox{0ex}[0ex][0ex]{\ensuremath{\hookleftarrow}},
}

\title{Adaptive Retrieval helps Reasoning in LLMs - but mostly if it's not used}
\workshoptitle{Workshop on Principles of Generative Modeling (PriGM)}

\author{
  Srijan Shakya$^{1,2}$, 
  Anamaria-Roberta Hartl$^{1}$, 
  Sepp Hochreiter$^{1,3}$, 
  Korbinian P\"oppel$^{4}$ \\[1em] 
  $^{1}$ Institute of Machine Learning, Johannes Kepler University Linz, Austria \\
  $^{2}$ Pro2future GmbH, Linz, Austria \\
  $^{3}$ NXAI GmbH, Linz, Austria \\ 
  $^{4}$ ELLIS Institute T\"ubingen, Germany \\
  \texttt{\{k12342997, hartl, hochreit, poeppel\}@ml.jku.at}
}

\begin{document}

\maketitle

\begin{abstract}
Large Language Models (LLMs) often falter in complex reasoning tasks due to their static, parametric knowledge, leading to hallucinations and poor performance in specialized domains like mathematics. This work explores a fundamental principle for enhancing generative models: treating retrieval as a form of dynamic in-context learning. We test an adaptive retrieval-augmented architecture where an LLM agent actively decides when to query an external knowledge base during its reasoning process. We compare this adaptive strategy against a standard Chain-of-Thought (CoT) baseline and a static retrieval approach on the GSM8K and MATH-500 benchmarks.
Although our experiments show that static retrieval is inferior to CoT, the adaptive retrieval shows interesting behavior: While traces \textit{including} retrieved results show slightly worse performance compared to CoT, traces that do not include retrieval actually perform better compared to CoT. This suggests that: (a) retrieval only rarely helps reasoning (we show a few counterexamples, e.g. using useful theorems) and (b) actively not using retrieval is indicative of good model performance.
Furthermore, we find that the model scales its retrieval frequency with the difficulty of the problem, reinforcing that the decision to retrieve is a crucial metacognitive signal. The agent's ability to self-assess its knowledge and selectively engage with external information represents a key principle for building more robust and reliable generative models.
\end{abstract}

\section{Introduction}
Recent breakthroughs in generative AI have revealed emergent capabilities like in-context learning that challenge classical theoretical frameworks \citep{brown2020languagemodelsfewshotlearners}. However, the knowledge encoded in Large Language Models (LLMs) is static, limiting their reliability in domains requiring up-to-date information or deep, specialized knowledge. This limitation is particularly acute in reasoning, where models often hallucinate or fail to execute precise, multi-step logic \citep{cobbe2021gsm8k}.

Retrieval-Augmented Generation (RAG) has emerged as a promising solution to ground LLM outputs in external knowledge \citep{lewis2021retrievalaugmentedgenerationknowledgeintensivenlp}. Yet, canonical RAG is often static: information is retrieved once based on an initial query and prepended to the context. This "retrieve-then-read" paradigm is sub-optimal for complex, multi-step reasoning problems where the need for specific information only emerges mid-way through the process \citep{trivedi-etal-2023-interleaving}.

This paper investigates a core principle for advancing generative models: defining retrieval as a form of dynamic in-context learning. We propose an adaptive retrieval framework where the LLM functions as an agent, capable of interleaving its chain-of-thought (CoT) reasoning with self-initiated calls to a tool~\citep{schick_toolformer_2023}. This allows the model to identify its own knowledge gaps and fetch relevant information precisely when needed. We hypothesize that this adaptive, on-demand approach will significantly outperform both a non-retrieval baseline and a static retrieval implementation. We test this on challenging mathematical reasoning benchmarks, providing empirical evidence for a principle that moves beyond static context to a more dynamic and effective form of knowledge integration.

\section{Method: An Adaptive Retrieval-Augmented Reasoning Agent}

Our system is built around an LLM agent that integrates reasoning with on-demand retrieval. The architecture consists of a core language model, a retrieval module, and a control mechanism governed by a specialized prompt.
\paragraph{Core Language Model} We use Llama-3.1-8B-Instruct as the reasoning agent in a zero-shot setting, chosen for its instruction-following capabilities and guided only by its system prompt.
\paragraph{Retrieval Module} We use a two-stage retrieval pipeline. In the first stage, a `BAAI/bge-m3` bi-encoder~\citep{chen_m3-embedding_2024} is used to generate dense embeddings for documents and queries, indexed with a FAISS HNSW~\citep{douze2024faiss, johnson2019billion} structure for efficient search. We experiment with two knowledge corpora: MathPile, a broad mathematical text collection \citep{wang2024mathpilebilliontokenscalepretrainingcorpus}, and OpenMathInstruct-2, a curated dataset of math question-answer pairs \citep{toshniwal2024openmath2}. In the second stage, the top candidates are re-ranked using the `BAAI/bge-m3-reranker` cross-encoder to ensure high relevance~\citep{li2023making,bge-m3}.

\section{Experiments and Results}

\paragraph{Reasoning Strategies} We compare three distinct strategies to isolate the impact of adaptive retrieval:
\begin{enumerate}
    \item \textbf{Baseline: Chain-of-Thought (CoT).} The LLM solves problems using only its internal knowledge, prompted to "think step-by-step."
    \item \textbf{Static Retrieval-Augmented-CoT.} We run a single retrieval once, using the original problem as the query then select the top-$k$ results and prepend them to the prompt, and then let the model perform CoT. There is no further retrieval during reasoning in this strategy.
    \item \textbf{Adaptive Retrieval-Augmented-CoT.} The LLM is instructed to use a special \texttt{<search>query</search>} tool whenever it identifies a knowledge gap during its reasoning. When this tag is generated, our system pauses generation, executes a search with the specified query, and injects the retrieved information back into the context. The LLM then continues its reasoning, now informed by the new evidence. This loop can be repeated until the model generates a final answer within \texttt{<answer>value</answer>} tags.
\end{enumerate}

\begin{figure}[h]
    \centering
    \includegraphics[width=1\linewidth]{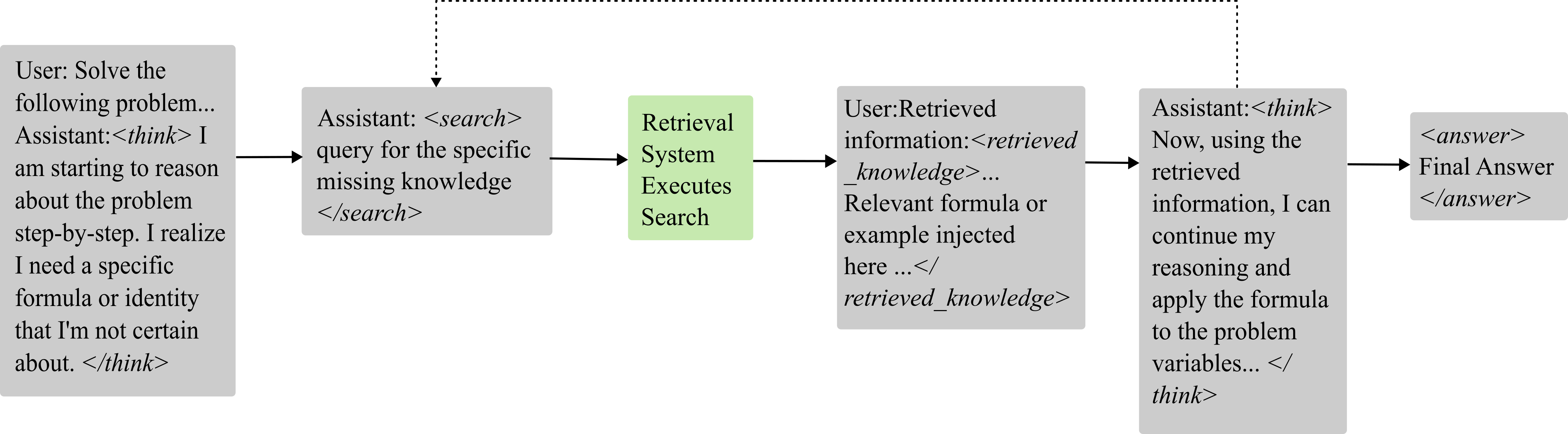}
    \caption{The dynamic context of the Adaptive Retrieval-CoT agent. Unlike a standard prompt, the context is an evolving transcript. The key difference is the agent's ability to generate a \texttt{<search>} tag, which pauses generation. The system then executes the query and injects the results back into the context, allowing the agent to resume its reasoning with new, targeted information.}
    \label{fig:placeholder}
\end{figure}

We evaluate the three strategies on the GSM8K \citep{cobbe2021gsm8k} and MATH-500 \citep{hendrycksmath2021} benchmarks, in a prompted zero shot setting. Exact prompts are shown in the appendix \ref{app:prompts}. The primary metric is the exact match accuracy.

\subsection{Overall Performance}
\begin{figure}[htbp]
    \includegraphics[width=\linewidth]{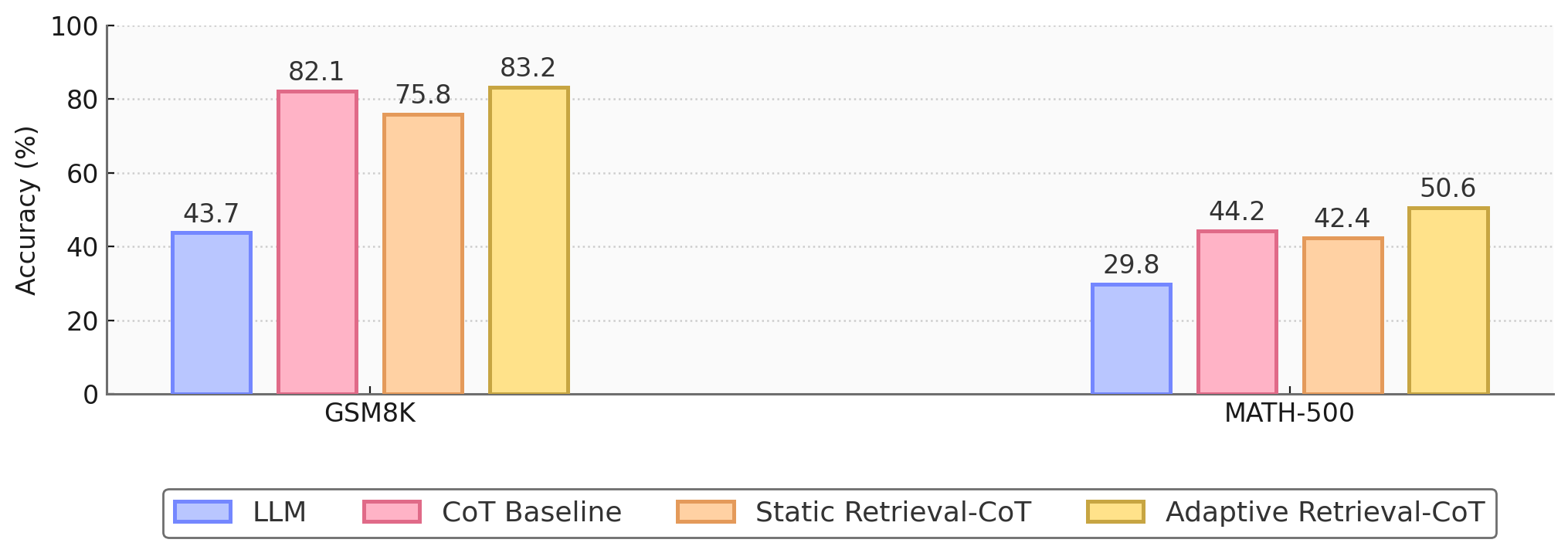}
  \caption{Performance comparison of the three reasoning strategies.}
  \label{fig:performance_chart}
\end{figure}

Table~\ref{tab:overall_accuracy} and Figure~\ref{fig:performance_chart} illustrate the main results. Adaptive Retrieval–CoT consistently achieves the highest accuracy, exceeding the CoT baseline by 1.1 percentage points(pp.) on GSM8K and by 6.4pp. on the more challenging MATH-500 dataset. These findings support our central hypothesis that enabling on-demand, agentic retrieval is an especially effective approach to enhancing mathematical reasoning.

Crucially, the Static Retrieval–CoT approach underperforms the CoT baseline. On GSM8K, its accuracy declines by 6.3pp., indicating that unsolicited, potentially noisy context provided a priori can disrupt the model’s reasoning. These results underscore that an adaptive agent’s ability to query information selectively is essential for effective knowledge integration.

\begin{table}[h]
  \caption{Overall accuracy comparison. Adaptive Retrieval-CoT significantly outperforms both the CoT baseline and the static RAG approach.}
  \label{tab:overall_accuracy}
  \centering
  \begin{tabular}{lccccc}
    \toprule
    Dataset   & LLM & \makecell{CoT \\ Baseline} & \makecell{Static \\Retrieval-CoT} & \makecell{Adaptive \\ Retrieval-CoT} & $\Delta$ vs CoT \\
    \midrule
    GSM8K     & 43.7\% & 82.1\%       & 75.8\%         & \textbf{83.2\%}  & +1.1pp          \\
    MATH-500  & 29.8\% & 44.2\%       & 42.4\%         & \textbf{50.6\%}  & +6.4pp          \\
    \bottomrule
  \end{tabular}
\end{table}

\vspace{-1em}

\vspace{-1em}

\subsection{Analysis of the Adaptive Agent's Behavior}

To understand the extent to which the agent relies on retrieval, we empirically analyze the agent's decision-making process. Accordingly, we identify a strong correlation between the problem complexity and the retrieval frequency.

On the GSM8K benchmark, the agent invoked retrieval in only \textbf{7.0\%} of cases, relying primarily on its parametric knowledge. By contrast, the agent used retrieval in 
\textbf{38.8\%} of cases on the much harder MATH-500 benchmark.

To further investigate the gap between the two benchmarks, we break down the MATH-500 dataset according to the difficulty levels (1–5; see Table~\ref{tab:difficulty_breakdown}).

The retrieval rate increases steadily with difficulty, from 14.0\% on Level 1 problems to 60.4\% on Level 5 problems. This demonstrates that the model correctly identifies more challenging problems as those requiring additional knowledge, a key component of effective dynamic in-context learning. 
The findings of this study indicate that utilizing the domain-specific OpenMathInstruct-2 corpus with a summarized injection format yields optimal results, suggesting that concise, high-quality knowledge is most advantageous.
\begin{table}[htbp]
  \caption{Accuracy and Retrieval Rate by difficulty level on the MATH-500 dataset. The model's retrieval usage scales with problem complexity.}
  \label{tab:difficulty_breakdown}
  \centering
  \begin{tabular}{lccccc}
    \toprule
    Difficulty Level & LLM & \makecell{CoT \\ Baseline} & \makecell{Static \\ Retrieval-CoT}   & \makecell{Adaptive \\ Retrieval-CoT}  & Retrieval Rate \\
    \midrule
    Level 1   & 51.2\% & 72.1\%       & 76.7\%         & \textbf{86.0\%}         & 14.0\%         \\
    Level 2  & 44.4\%  & 60.0\%       & 63.3\%         & \textbf{67.8\%}         & 21.1\%         \\
    Level 3   & 37.1\% & 52.4\%       & 49.5\%         & \textbf{61.9\%}         & 33.3\%         \\
    Level 4   & 21.9\% & 38.3\%       & 32.0\%         & \textbf{47.7\%}         & 41.4\%         \\
    Level 5  & 14.9\%  & \textbf{23.9\%}       & 21.6\%         & 21.6\%         & 60.4\%         \\
    \bottomrule
  \end{tabular}
\end{table}

\subsection{Retrieval Decision Analysis}
\label{app:retrieval-decision-analysis}

\begin{figure}[!htbp]
  \centering
  \begin{minipage}{0.497\textwidth}
    \centering
    \includegraphics[width=\linewidth]{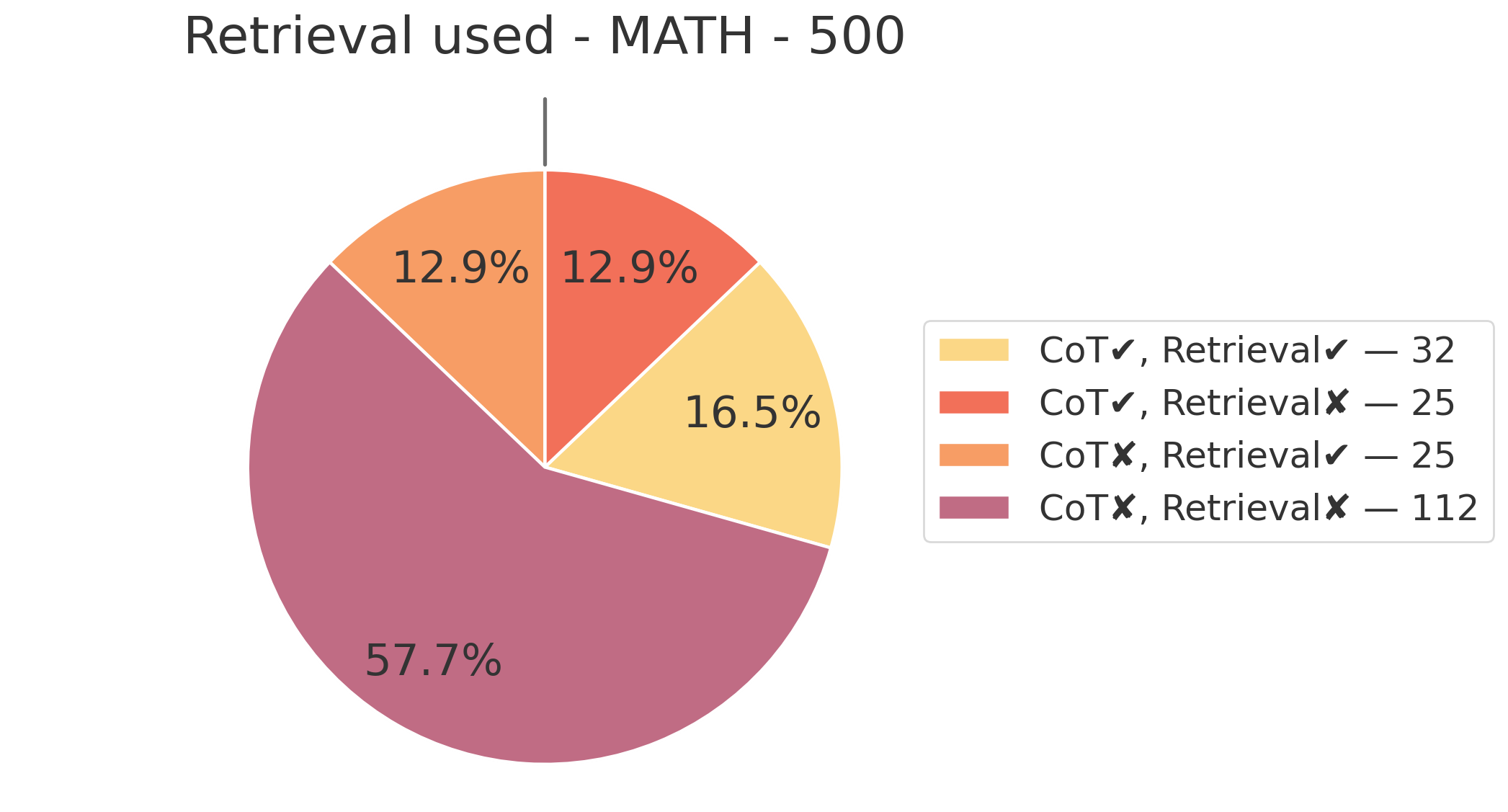}\\
  \end{minipage}
  \hfill
  \begin{minipage}{0.497\textwidth}
    \centering
    \includegraphics[width=\linewidth]{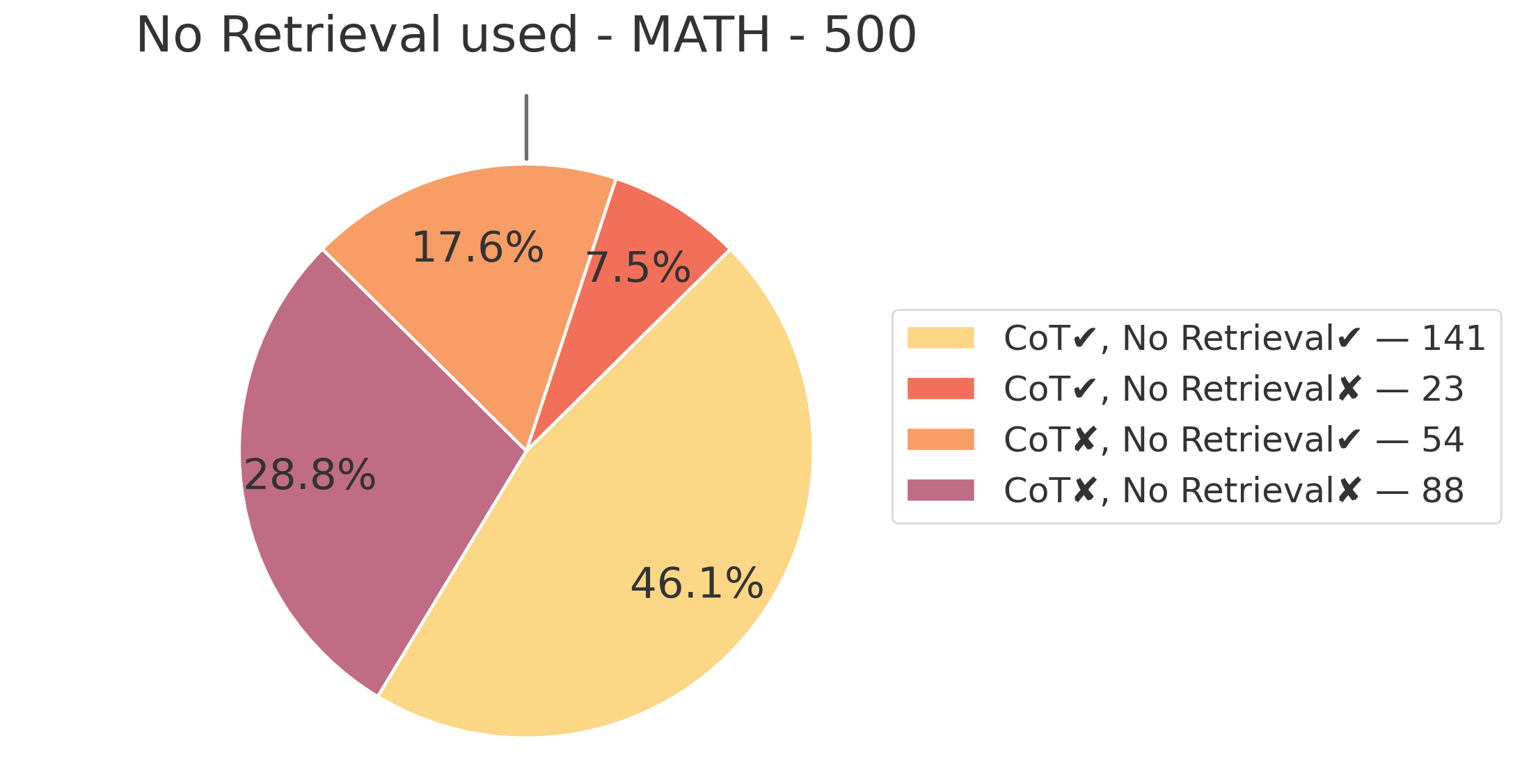}
  \end{minipage}
  \caption{Contingency: CoT vs Adaptive Retrieval-CoT on \textbf{MATH-500}, \ding{52} means correctly solved, \ding{56}  means the method didn't solve the task. Top line indicates decision boundary for better method.}
  \label{fig:side-by-side}
\end{figure}

To understand how retrieval decisions affect correctness, we compare outcomes of the CoT baseline and the Adaptive Retrieval-CoT model. The contingency charts below summarize the best-performing configurations for benchmark \textbf{MATH-500} - Figure \ref{fig:side-by-side}.

\begin{figure}[!htbp]
  \centering
  \begin{minipage}{0.49\textwidth}
    \centering
    \includegraphics[width=\linewidth]{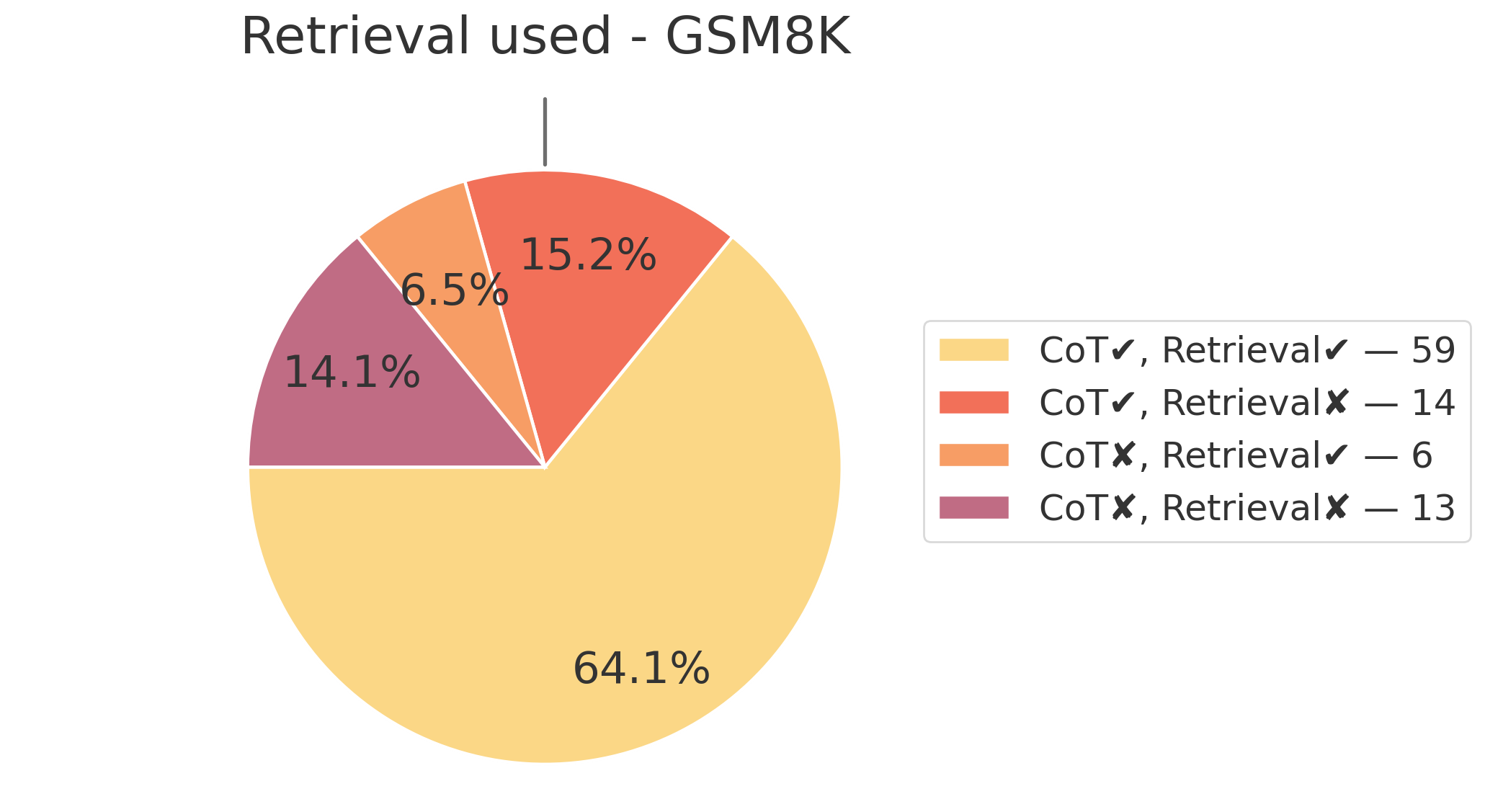}\\
  \end{minipage}
  \hfill
  \begin{minipage}{0.49\textwidth}
    \centering
    \includegraphics[width=\linewidth]{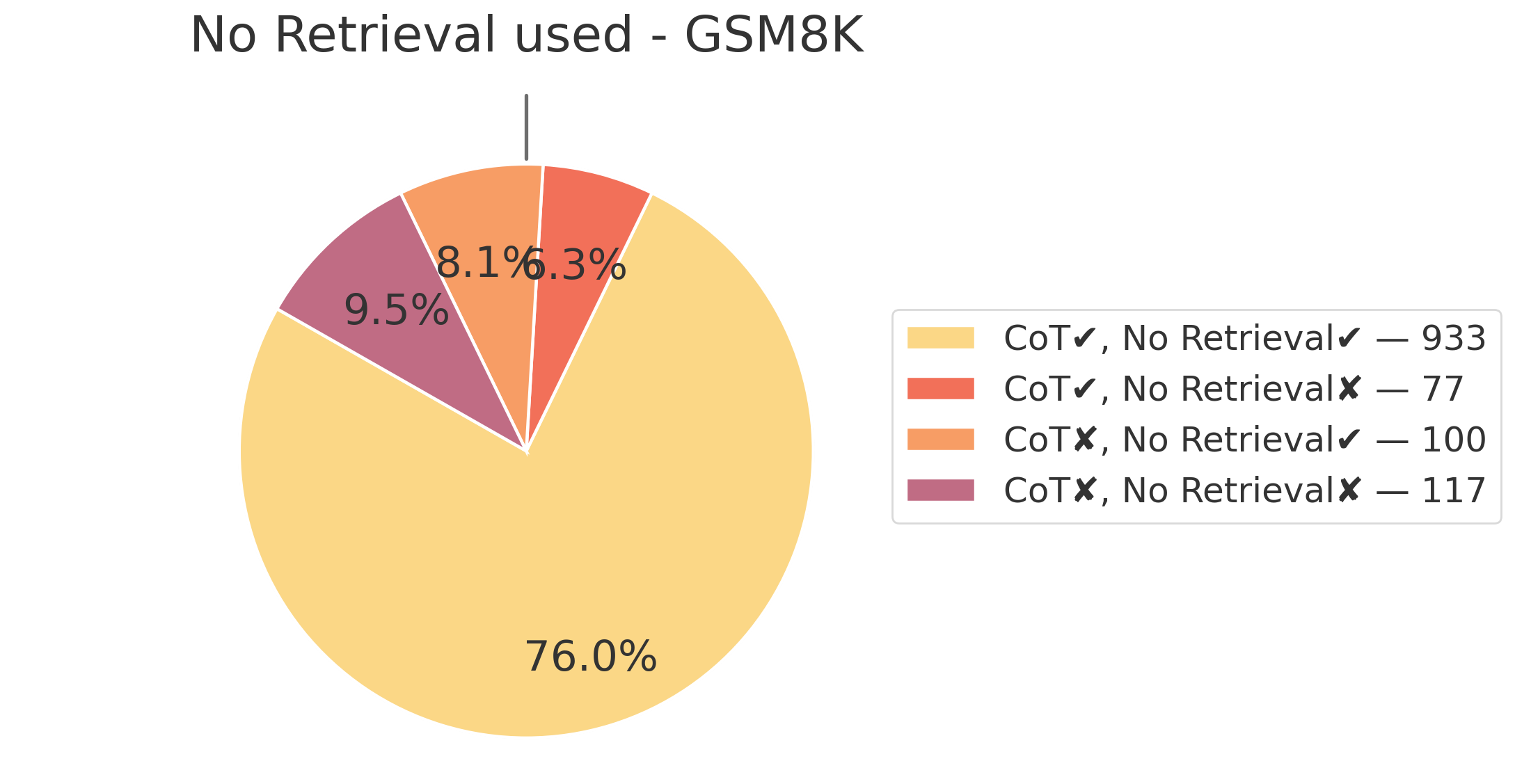}
  \end{minipage}
  \caption{Contingency: CoT vs Adaptive Retrieval-CoT on \textbf{GSM8K}, \ding{52} means correctly solved, \ding{56}  means the method didn't solve the task. Top line indicates decision boundary for better method.}
  \label{fig:side-by-side-gsm8k}
\end{figure}

\noindent
\textbf{Key observations.} 
On GSM8K, retrieval was invoked in only 7\% of problems (92/1319). 
When the baseline CoT was already correct, retrieval rarely improved performance—helping in 57 cases (5.3\%) and hurting in 16 (1.5\%) and when CoT was wrong, retrieval corrected only 6 cases (2.5\%). 
Despite this low retrieval precision (27.3\%), overall accuracy still rose from 82.1\% to 83.2\%, suggesting that the benefit stems from the \emph{retrieval-triggered reflection process} rather than the retrieved content itself.

On the MATH-500 benchmark, retrieval was used far more frequently (38.8\%) and demonstrated a balanced outcome: 25 helped and 25 hurt cases. 
Although retrieval itself was not consistently reliable, it was predominantly triggered on complex or uncertain problems, functioning as a form of \emph{uncertainty-aware reasoning}. 
The model effectively “knows when it doesn’t know,” invoking retrieval as a signal to pause and reconsider its reasoning path, which led to a net performance gain of +6.4~pp over the CoT baseline.

Perhaps the most striking observation is that the agent’s performance is highest when it \textbf{chooses not to retrieve}. In cases where the agent did not invoke a search, its accuracy was 84.2\% on GSM8K and 63.7\% on MATH-500—outperforming the standard CoT baseline by +2.1pp and a massive +19.5pp, respectively. This strongly suggests that the agent's decision to forgo retrieval is a reliable indicator of its confidence and correctness, reinforcing the idea that the adaptive framework's true benefit lies in the metacognitive ability to assess its own knowledge gaps.

\section{Discussion and Conclusion}

This work provides empirical grounding for a broader principle of generative modeling: \textbf{retrieval can act as a mechanism for dynamic in-context learning}. In contrast to static generative models that rely on fixed parametric knowledge, an adaptive retrieval agent treats the context as an evolving workspace that it can query, revise, and expand during reasoning. This reframes generation not as a one-shot mapping from prompt to output, but as an iterative process of \emph{context construction under uncertainty}.

Our results illustrate the practical and theoretical significance of this shift. First, retrieval is not uniformly beneficial; static context injection often harms performance by introducing irrelevant noise. However, when retrieval is made \emph{agentic}, when the model decides \emph{when} and \emph{what} to retrieve, performance improves consistently (+1.1pp on GSM8K, +6.4pp on MATH-500). This highlights a key insight: the value lies not just in the retrieved content but in the model's capacity to control its own informational boundaries. The agent's decision to retrieve scales with problem difficulty demonstrates an adaptive allocation of computational effort.

Most tellingly, the agent's performance is highest when it forgoes retrieval, suggesting its decision acts as a reliable signal of its own competence. The LLM effectively regulates the flow of external information, invoking retrieval as a marker of epistemic uncertainty rather than as a default operation. In this sense, retrieval becomes a form of \textbf{self-reflective computation}: a way for a generative model to acknowledge when it “does not know,” and to act upon that realization. This principle of self-regulated knowledge integration offers a promising path toward more robust and reliable generative AI.

\bibliographystyle{unsrt}
\bibliography{references}

\clearpage
\appendix
\section{Appendix}

\subsection{Experimental Setup and Hyperparameter Configuration}
\label{app:hyperparams}
The following table details the specific hyperparameters and configuration settings used for the different experimental runs. 

\begin{table}[h]
    \centering
    \caption{Key Hyperparameters and Configuration Settings.}
    \label{tab:exp_hyperparameters_detailed}
    \renewcommand{\arraystretch}{1.2} %
    \begin{tabular}{ll}
        \toprule
        \textbf{Parameter} & \textbf{Value} \\
        \midrule
        \multicolumn{2}{l}{\textit{Models \& Infrastructure}} \\
        Core Language Model & \texttt{meta-llama/Llama-3.1-8B-Instruct} \\
        Retrieval \& Re-ranking Model & \texttt{BAAI/bge-m3} \\
        Model Precision & \texttt{float16} \\
        Vector Index Type & FAISS (HNSW) \\
        \midrule
        \multicolumn{2}{l}{\textit{Generation Configuration}} \\
        Reasoning Temperature & 0.0 (Greedy Decoding) \\
        Max New Tokens & 1024 \\
        \midrule
        \multicolumn{2}{l}{\textit{Retrieval Configuration}} \\
        Initial Candidates Retrieved (\texttt{k\_dense}) & 200 \\
        Final Documents after Re-rank (\texttt{k\_final}) & 5 \\
        Re-ranking Method & \texttt{bge-m3} ColBERT scores \\
        \bottomrule
    \end{tabular}
\end{table}

\subsection{Prompt Templates}
\label{app:prompts}
This appendix lists the exact prompts used in our experiments. We report them verbatim to support replication.

\subsubsection{Chat Formatting (Llama 3 Template)}
\label{app:llama3-template}

\begin{Prompt}
<|begin_of_text|><|start_header_id|>system<|end_header_id|>{system}<|eot_id|>
<|start_header_id|>user<|end_header_id|>{user}<|eot_id|>
<|start_header_id|>assistant<|end_header_id|>
\end{Prompt}

\subsubsection{LLM with No CoT (System Prompt)}
\label{app:prompts-llm}
\begin{Prompt}
You are an expert mathematician.
Solve the math problem and produce **exactly one**

    <answer> your final answer </answer>
    
Nothing after `</answer>`.
\end{Prompt}

\subsubsection{Zero-Shot CoT (System Prompt)}
\label{app:prompts-cot}
\begin{Prompt}
You are an expert mathematician.

Think step-by-step.  

Write every reasoning step inside '<think> ... </think>' blocks. 

When you are completely done, produce **exactly one**

    <answer> your final answer </answer>
    
Nothing after `</answer>`.
\end{Prompt}

\subsubsection{Static Retrieval-CoT (System Prompt)}
\label{app:prompts-static}
\begin{Prompt}
You are an expert mathematician.

Combine the retrieved information below with your own knowledge to solve the problem.Use the context mainly as examples or hints—you may add any facts you already know.

Think step-by-step.
Write every reasoning step inside `<think> … </think>` blocks.

When you are completely done, produce **exactly one**
    <answer> your final answer </answer>
Note: Nothing after `</answer>`.
\end{Prompt}

\subsubsection{Adaptive Retrieval-CoT (System Prompt)}
\label{app:prompts-adaptive}
\begin{Prompt}
You are an expert mathematician.
Think step-by-step.
Write every reasoning step inside `<think> … </think>` blocks.


If you need to look up a formula, definition, or problems, you can use the <search> tool by writing a search query inside the <search> tag like this: 
<search>your search query</search>

After retrieval, you may:
1. Use the information if helpful
2. Explicitly state "Retrieved information not helpful" and continue without it

After the search results are returned, continue your step-by-step thinking.

When you are ready to give the final answer, use the <answer> tag like 
this: <answer>your final answer</answer>
\end{Prompt}

\subsubsection{Integration Prompt (Used After Retrieval)}
\label{app:prompts-adaptive-integration}
\begin{Prompt}
Retrieved information:
<retrieved_knowledge>
{injected_text}
</retrieved_knowledge>

Strict rules for integration:
1. Extract the core mathematical principle (e.g., formula or theorem) in abstract terms—ignore specific numbers or examples.
2. State: 'Applying [principle name]: [abstract formula]'.
3. Map this to your problem's variables and show step-by-step application.
4. If the retrieval is not helpful (e.g., irrelevant or too specific), explicitly state 'Retrieval not used' and continue with your original reasoning.
5. Never copy numbers or solutions—adapt abstractly to avoid errors.

\end{Prompt}

\subsubsection{Summary Prompt (Canonical Method Extraction)}
\label{app:prompts-adaptive-summary}
\begin{Prompt}
From the following evidence, extract a single, reusable method (definition, theorem, or algorithm) in a canonical, abstract form.

Your output must be a single, concise sentence following these rules:
- **Structure:** Start with the method's canonical name, followed by its formula in abstract variables, and end with any critical preconditions.
- **Content:** The formula must use abstract variables (e.g., a, b, n, x) not numbers from the problem.
- **Format:** Output only the single sentence. If no single canonical method can be extracted, output "UNHELPFUL".

**Good Example:**
The Binomial Theorem states that for a positive integer n, $(x+a)^n = \sum_{k=0}^{n} \binom{n}{k} x^k a^{n-k}$.

**Bad Example:**
To solve this, we used the binomial theorem to expand (2x+3)^4.
\end{Prompt}

\subsection{Full Example Reasoning Traces}
\label{app:example-reasoning-traces}

\subsubsection{Example where CoT Failed but Adaptive Retrieval-CoT succeeded}
\textbf{Problem.} Let
\[
x^8 + 3x^4 - 4 \;=\; p_1(x)\,p_2(x)\,\dotsm\,p_k(x),
\]
where each non-constant $p_i(x)$ is monic with integer coefficients and irreducible over $\mathbb{Z}$. Compute $p_1(1) + \cdots + p_k(1)$.

\smallskip
\noindent\textbf{CoT (wrong, no retrieval).}
\begin{quote}\small\ttfamily
<think>
We want to factor $x^8 + 3x^4 - 4$ into irreducibles over the integers and then sum $p_i(1)$.\\
Try: $x^8 + 3x^4 - 4 \;=\; (x^4 + 4)(x^4 - 1)$.\\
Further: $x^4 - 1 \;=\; (x^2 + 1)(x^2 - 1) \;=\; (x^2 + 1)(x+1)(x-1)$.\\
Assume $x^4 + 4$ is irreducible over $\mathbb{Z}$. Then the factors are $(x^4 + 4)$, $(x^2 + 1)$, $(x+1)$, $(x-1)$.\\
Evaluate at $x{=}1$: $5 + 2 + 2 + 0 \;=\; 9$.\\
</think>
\texttt{<answer>9</answer>}
\end{quote}

\noindent\textbf{Adaptive Retrieval-CoT (correct, with retrieval).}
\begin{quote}\small\ttfamily
<think>
We factor and then sum $p_i(1)$. Start as before: $x^8 + 3x^4 - 4 \;=\; (x^4 + 4)(x^4 - 1)$ and $x^4 - 1 \;=\; (x^2 + 1)(x-1)(x+1)$.\\
Key question: is $x^4 + 4$ reducible over $\mathbb{Z}$?\\
</think>
<search>factor $x^4 + 4$ over integers (Sophie Germain)</search>\\
\textit{Retrieved:} $x^4 + 4 \;=\; x^4 + 4x^2 + 4 - 4x^2 \;=\; (x^2 + 2x + 2)(x^2 - 2x + 2)$, irreducible quadratics over $\mathbb{Z}$.\\
<think>
So the full factorization over $\mathbb{Z}$ is\\
$(x^2 + 2x + 2)(x^2 - 2x + 2)(x^2 + 1)(x-1)(x+1)$.\\
Evaluate at $x{=}1$: $(1+2+2) + (1-2+2) + (1+1) + (1-1) + (1+1) \;=\; 5 + 1 + 2 + 0 + 2 \;=\; 10$.\\
</think>
\texttt{<answer>10</answer>}
\end{quote}

\noindent\textbf{Query used:} \texttt{factor $x^4 + 4$ over integers (Sophie Germain)} \quad
\textbf{Retrieved:} \texttt{$x^4+4=(x^2+2x+2)(x^2-2x+2)$}
\paragraph{Comment.}
The baseline CoT fails by implicitly assuming $x^4+4$ is irreducible over $\mathbb{Z}$, yielding an incorrect sum of $9$. 
The adaptive Retrieval–CoT run issues a single, targeted lookup to confirm the Sophie–Germain identity $x^4+4=(x^2+2x+2)(x^2-2x+2)$ and then completes the factorization $(x^2+2x+2)(x^2-2x+2)(x^2+1)(x-1)(x+1)$, giving $p_1(1)+\cdots+p_k(1)=10$. 
This illustrates retrieval as \emph{on-demand memory}: a sparse, uncertainty-triggered intervention that supplies a specific missing identity, rather than indiscriminate context stuffing. 
The gain comes from \emph{when} retrieval is used and \emph{what} is retrieved—supporting our claim that timing and agency, not volume of context, drive the benefits of adaptive retrieval.

\subsubsection{Example where CoT succeeded but Adaptive Retrieval-CoT Failed}
\textbf{Problem.} Melanie is a door-to-door saleswoman. She sold a third of her vacuum cleaners at the green house, $2$ more to the red house, and half of what was left at the orange house. If Melanie has $5$ vacuum cleaners left, how many did she start with?

\smallskip
\noindent\textbf{CoT (correct, no retrieval).}
\begin{quote}\small
\textbf{\texttt{<think>}}\\
Let the starting total be $x$.\\
After green house: sold $x/3$, kept $x - x/3 = 2x/3$.\\
After red house: kept $2x/3 - 2$.\\
After orange house: sold half, so kept $\tfrac{1}{2}\bigl(2x/3 - 2\bigr) = x/3 - 1$.\\
Set remaining to $5$: $x/3 - 1 = 5 \Rightarrow x/3 = 6 \Rightarrow x = 18$.\\[2pt]
\textbf{\texttt{</think>}} \quad \texttt{<answer>}18\texttt{</answer>}
\end{quote}

\noindent\textbf{Adaptive Retrieval-CoT (wrong, retrieval hurt).}
\begin{quote}\small
\textbf{\texttt{<think>}}\\
Let the starting total be $x$. After green: $2x/3$. After red: $2x/3 - 2$.\\
After orange: “half of what was left” interpreted as subtracting $\tfrac{1}{2}(2x/3 - 2)$: \\
\phantom{After orange: } remaining $= 2x/3 - 2 - \bigl(x/6 - 1\bigr) = x/2 - 1 \stackrel{?}{=} 5 \Rightarrow x = 12$.\\
\textbf{\texttt{</think>}}\\[2pt]
\texttt{<search>}reverse-operations fraction word problems method\texttt{</search>}\\
\emph{Retrieved (generic tip):} “For multi-step fraction sales, consider reversing from the final amount.”\\[2pt]
\textbf{\texttt{<think>}}\\
Apply reverse steps in a different order $\Rightarrow$ inconsistent checks; keep $x=12$.\\[2pt]
\textbf{\texttt{</think>}} \quad \texttt{<answer>}12\texttt{</answer>}
\end{quote}

\paragraph{Comment.} The phrase “half of what was left” means the remaining stock is halved (i.e., keep $\tfrac{1}{2}$ of the current amount). The CoT sets remaining to $x/3 - 1$ and solves $x=18$. The dynamic run’s mid-trace retrieval introduces a generic heuristic that derails a correct plan, yielding $12$. This is a typical case where \emph{unsolicited or mistimed retrieval} reduces coherence on an otherwise routine backtracking problem.

\subsubsection{Example where CoT Failed but Adaptive Retrieval-CoT succeeded without use of Retrieval}

\textbf{Problem.} Kylar went to the store to buy glasses for his new apartment. One glass costs \$5, but every second glass costs only \(60\%\) of the price. Kylar wants to buy \(16\) glasses. How much does he need to pay for them?

\smallskip
\noindent\textbf{CoT (wrong, no retrieval).}
\begin{quote}\small
\textbf{\texttt{<think>}}\\
Treat first at full price and the remaining at discount: first costs \$\(5\). Remaining \(16-1=15\) glasses at \(0.6\times 5=\$3\) each.\\
Total \(= 5 + 15\cdot 3 = 5 + 45 = \$50\).\\[2pt]
\textbf{\texttt{</think>}} \quad \texttt{<answer>}50\texttt{</answer>}
\end{quote}

\noindent\textbf{Adaptive Retrieval-CoT (correct, \emph{no} retrieval).}
\begin{quote}\small
\textbf{\texttt{<think>}}\\
“Every second glass” means prices alternate by pairs: for each \emph{pair}, first is full price, second is \(60\%\) of full.\\
Full price \(=\$5\), discounted price \(=0.6\times 5=\$3\).\\
Cost per pair \(= 5 + 3 = \$8\). With \(16\) glasses we have \(16/2 = 8\) pairs.\\
Total \(= 8 \times 8 = \$64\).\\[2pt]
\textbf{\texttt{</think>}} \quad \texttt{<answer>}64\texttt{</answer>}
\end{quote}

\paragraph{Comment.} The CoT misreads the pattern and applies the discount to all but the first glass. The adaptive run (still without retrieval) recognizes the alternating-price structure and computes by pairs, yielding the correct total \(\$64\).

\newpage

\subsection{Detailed Contingency Analysis}
\label{app:contingency_tables}

The following tables provide a detailed breakdown of outcomes for the Adaptive Retrieval-CoT agent compared to the CoT baseline on the GSM8K and MATH-500 benchmarks. These tables contain the raw data visualized in the main text and form the basis for the key observations in the \ref{app:retrieval-decision-analysis} section.

\begin{table}[!htbp]
\centering
\footnotesize
\caption{Contingency table: CoT vs.\ Adaptive Retrieval-CoT on \textbf{GSM8K}.}
\label{tab:gsm8k_contingency}
\vspace{0.4em}
\begin{tabular}{lccccc}
\toprule
 & \makecell{Retrieval \\ + Correct} 
 & \makecell{Retrieval \\ + Incorrect} 
 & \makecell{No Retrieval \\ + Correct} 
 & \makecell{No Retrieval \\ + Incorrect} 
 & Total \\
\midrule
CoT Correct & 59 & 14 & 933 & 77 & 1083 \\
CoT Incorrect & 6 & 13 & 100 & 117 & 236 \\
\midrule
Total & 65 & 27 & 1033 & 194 & 1319 \\
\bottomrule
\end{tabular}
\end{table}

\begin{table}[!htbp]
\centering
\footnotesize
\caption{Contingency table: CoT vs.\ Adaptive Retrieval-CoT on \textbf{MATH-500}.}
\label{tab:math500_contingency}
\vspace{0.4em}
\begin{tabular}{lccccc}
\toprule
 & \makecell{Retrieval \\ + Correct} 
 & \makecell{Retrieval \\ + Incorrect} 
 & \makecell{No Retrieval \\ + Correct} 
 & \makecell{No Retrieval \\ + Incorrect} 
 & Total \\
\midrule
CoT Correct & 32 & 25 & 141 & 23 & 221 \\
CoT Incorrect & 25 & 112 & 54 & 88 & 279 \\
\midrule
Total & 57 & 137 & 195 & 111 & 500 \\
\bottomrule
\end{tabular}
\end{table}

\newpage

\end{document}